\documentclass[letterpaper, 10 pt, conference]{ieeeconf}  
\usepackage{amsmath}
\usepackage{booktabs}
\usepackage{amsmath}
\usepackage{dsfont}
\usepackage{amssymb}
\usepackage{graphicx}
\usepackage[table]{xcolor}

\IEEEoverridecommandlockouts                              

\usepackage{graphicx}
\usepackage{etoolbox}
\usepackage{capt-of}
\usepackage{url}          
\usepackage{hyperref}
\makeatletter
\renewcommand\@makefntext[1]{\noindent #1}
\makeatother

\makeatletter
\apptocmd{\@maketitle}{%
  \vspace{6pt}
  \centering
  \includegraphics[width=\textwidth]{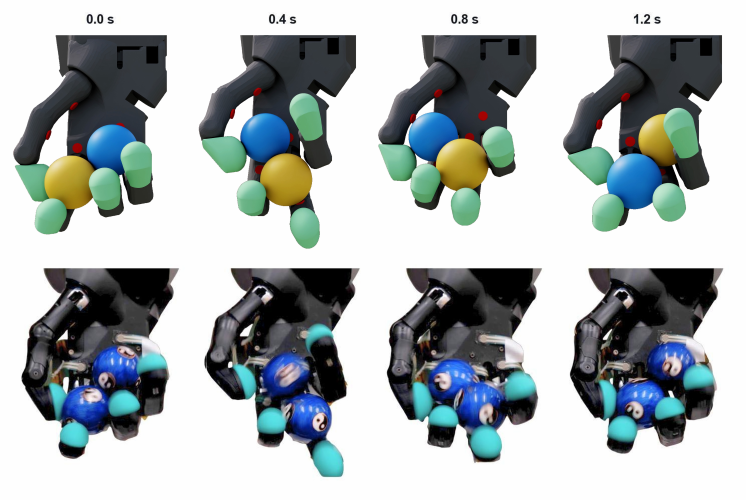}
  \captionof{figure}{\textbf{Zero-shot sim-to-real transfer of blind dexterous manipulation.} 
A 1.2-second $180^\circ$ Baoding ball rotation sequence in simulation (\textbf{top}) and on the physical Shadow Dexterous Hand Lite (\textbf{bottom}) at matched time steps. 
Trained entirely in simulation and deployed zero-shot, the policy operates using only proprioception and binary tactile contact signals, without vision, depth, motion capture, or object-state estimation entering the pipeline. 
Under this blind setup, the real-world policy completes \textbf{225 consecutive $180^\circ$ rotations} (112.5 full rotations) in a single continuous trial without dropping a ball.}
  \label{fig:final-tactile-Baoding}
  \vspace{6pt}
}{}{\PackageError{}{patching \string\@maketitle\space failed}{}}

\makeatother

\title{\LARGE \bf
No Eyes, No Problem: What Does Touch Contribute to Blind Baoding Ball Manipulation?
}

\author{Ayush Deshmukh$^{1,*}$, Nalin Agarwal$^{1,*}$, Elle Miller$^{1}$, Sethu Vijayakumar$^{1}$
\thanks{$^{*}$Equal contribution.}%
\thanks{$^{1}$All authors are with the School of Informatics, University of Edinburgh, Edinburgh, UK.
        {\tt\small \{a.deshmukh, n.agarwal-10\}@sms.ed.ac.uk,}
        {\tt\small \{elle.miller, sethu.vijayakumar\}@ed.ac.uk}}%
}

\begin{document}

\maketitle

\setcounter{figure}{1}
\begin{abstract}

Humans manipulate objects within their hands through touch alone with remarkable ease, but for robots, achieving sustained real-world manipulation without vision remains an open challenge. Although reinforcement learning (RL) can generate sophisticated blind manipulation policies in simulation, transferring them to a real dexterous hand remains hindered by continuously changing contacts, motion-induced tactile information drift, and the gap between simulated and physical dynamics. We study this problem through Baoding ball rotation on the Shadow Dexterous Hand Lite, using only proprioception and binary tactile feedback as observations. We achieve successful zero-shot transfer through a combination of a sensor-aligned, coupling-aware embodiment in simulation, tactile calibration, and two novel episodic domain randomisations: slew-rate action limits and STuck-AT Taxels. On the physical hand, the policy achieves sustained blind Baoding rotation, completing 225 consecutive $\mathbf{180^\circ}$ rotations (112.5 full rotations) at $\sim0.36 \;\text{rot/s}$ in a single continuous trial without any vision or object-state information. Furthermore, we evaluate the contribution of tactile sensing relative to proprioception alone and identify the hand regions where tactile information is most important. This is the strongest reported hardware result for Baoding ball manipulation in both rotation count and speed, irrespective of the sensing modality used. By effectively bridging the sim-to-real gap, our framework sets a new benchmark for sustained tactile-driven dexterous manipulation on physical hardware. Videos and additional results are available on the project website: \url{https://noeyesnoproblem.github.io/}.



\end{abstract}

\section{INTRODUCTION}
\label{sec:intro}

In-hand manipulation requires continuous control of contact, with fingers making, breaking, and re-establishing contact as the object moves relative to the hand. The hand is simultaneously the actuator and the sensor, so each corrective motion changes the contact state from which the next correction is made. This intermittent contact and discontinuous dynamics make in-hand manipulation difficult to model and control. Most approaches address this challenge by supplying object state externally, through cameras, markers, or privileged simulator information. These observations provide explicit information about object pose and geometry, yet humans can reorient objects in the hand without continuously observing them, relying instead on contact during manipulation.
This raises a fundamental question: \textit{can contact alone stand in for explicit object state, as it does for humans?}

We take this question literally. Unlike prior systems that rely on
cameras~\cite{handa2023dextreme, chen2023visual},
motion-capture markers~\cite{andrychowicz2020learning}, or privileged
simulator state distilled into a student
policy~\cite{chen2022system, qi2023inhand}, our policy receives none
of these; it observes only joint encoders and 16 binary contact
signals, a representation shown to be sufficient in
simulation for single-object
rotation on hardware~\cite{yin2023rotating}.
While most existing real-world in-hand manipulation studies focus on reorienting a single object via reinforcement learning (RL) \cite{andrychowicz2020learning, sievers2022learning, handa2023dextreme, yin2023rotating, chen2023visual, qi2023inhand}, we choose to study Baoding ball manipulation as a more challenging instance of in-hand dexterity: two objects circulate simultaneously, interact with each other as well as the fingers, and are manipulated here on only a four-fingered Shadow Dexterous Hand Lite. The phase relationship between the two objects must be maintained without being directly observed. 

Prior approaches that attempt Baoding ball rotation have typically used external object-state information or tactile sensing to demonstrate rotation~\cite{nagabandi2020deep, yuan2024robot, yang2023tacgnn}. However, sustained zero-shot transfer of a blind policy from simulation to physical hardware remains an open challenge.
More recently, dexterous Baoding ball rotation was demonstrated in simulation across multiple robotic hands using only proprioception and binary contacts \cite{miller2026roto}. It was shown that combining RL with self-supervision (e.g. latent multi-step forward dynamics) learned implicit representations of the manipulated object(s) state, which led to superhuman simulated blind dexterity \cite{miller2026enhancing}. Yet, these works did not provide any hardware results, so the transferability of these results was unknown. 

In this work, we achieve zero-shot sim-to-real transfer of Baoding ball rotation using only proprioceptive and binary contacts as observations, extending upon the simulation results in \cite{miller2026enhancing, miller2026roto}. We construct a simulation embodiment that reproduces the physical tendon coupling and tactile sensor layout used on the Shadow Dexterous Hand Lite, and introduce per-channel tactile calibration, command slew-rate randomisation and Stuck-AT Taxels (STAT) to improve robustness to hardware discrepancies. We show that a policy trained entirely in simulation can complete 225 consecutive $180^\circ$ rotations (112.5 full rotations) at $\sim0.36 \;\text{rot/s}$ in a single continuous trial on a physical Shadow Hand Lite  from touch and proprioception alone. Our main contributions are as follows:

\begin{itemize}
\item A hardware-aligned Shadow Dexterous Hand Lite simulation with matched tactile sensor sites and action mapping that reproduces the physical tendon coupling.
\item A per-channel hysteresis-based tactile calibration to mitigate sensor noise.
\item Two episode-level randomisations, Slew-rate Domain Randomisation (SlewDR) and Stuck-AT Taxels (STAT), that encourage smoother policy outputs and reduce reliance on individual tactile channels.
\item Rigorous benchmarking of Baoding ball rotation across increasing levels of feedback information and cumulative randomisation ablations.
\item Zero-shot transfer of a blind Baoding policy to the Shadow Dexterous Hand Lite, sustaining rotation from proprioception and binary contact, with tactile feedback substantially improving performance and long-horizon recovery.
\end{itemize}

\section{Related Works}
\label{sec:related}

\textbf{In-hand manipulation} involves contacts that continuously form, break, and shift
as the object moves relative to the hand. RL in simulation
with domain randomization has made such skills practical, from block
reorientation~\cite{andrychowicz2020learning} and Rubik's cube
manipulation~\cite{akkaya2019solving} on the Shadow Hand to agile cube
reorientation on the Allegro Hand~\cite{handa2023dextreme}. Later work
generalized across object shapes and sensing modalities, using object pose and
point clouds~\cite{chen2022system}, a single depth camera~\cite{chen2023visual},
or vision combined with dense touch~\cite{qi2023general}. These methods obtain
object state from outside the hand, even though the object is often occluded
once the fingers enclose it.

\textbf{Blind in-hand manipulation} relies on the hand's own sensing rather
than external perception. Proprioception has
enabled single-object rotation~\cite{qi2023inhand}, and joint position and
torque sensing has enabled sustained cube rotation and
reorientation~\cite{sievers2022learning,pitz2023dextrous}. However, light objects
produce only small changes in the hand state \cite{miller2026enhancing}, which motivates direct contact
sensing. Dense and optical tactile representations provide richer contact
information but are harder to simulate and
calibrate~\cite{yang2024anyrotate,pan2026beyond}, whereas binary contacts are
compact and transfer more readily~\cite{yin2023rotating,lee2024dextouch}. Existing blind hardware results with sustained manipulation involve a single object.

\textbf{Baoding ball manipulation} extends blind in-hand manipulation to the coordination of multiple free objects, requiring two balls to be coordinated through both ball--hand and ball--ball contacts, making it more challenging than single-object manipulation. Prior
hardware results (Table~\ref{tab:main_results}) rely largely on external object
information. Nagabandi et al.~\cite{nagabandi2020deep} used stereo vision with model
predictive control on a Shadow Hand, while Yuan et al.~\cite{yuan2024robot} fused
depth point clouds with binary FSR sensing on an Allegro Hand. Blind hardware
evidence remains limited. Yang et al.~\cite{yang2023tacgnn} estimated ball positions
from a 653-taxel tactile array using a graph network, but reported only a single
qualitative half-rotation on hardware.
RoTO~\cite{miller2026enhancing} and RoTO~2.0~\cite{miller2026roto} used binary tactile sensing,
including on the Shadow Dexterous Hand Lite, but remain in simulation. Moreover, none of these hardware works demonstrates rapid rotation sustained
over several minutes. We address this gap through zero-shot transfer of a blind policy to the
Shadow Dexterous Hand Lite, using only proprioception and binary tactile
sensing, and show sustained rotation on hardware.

\begin{figure}[t]
    \centering
\includegraphics[width=1\columnwidth]{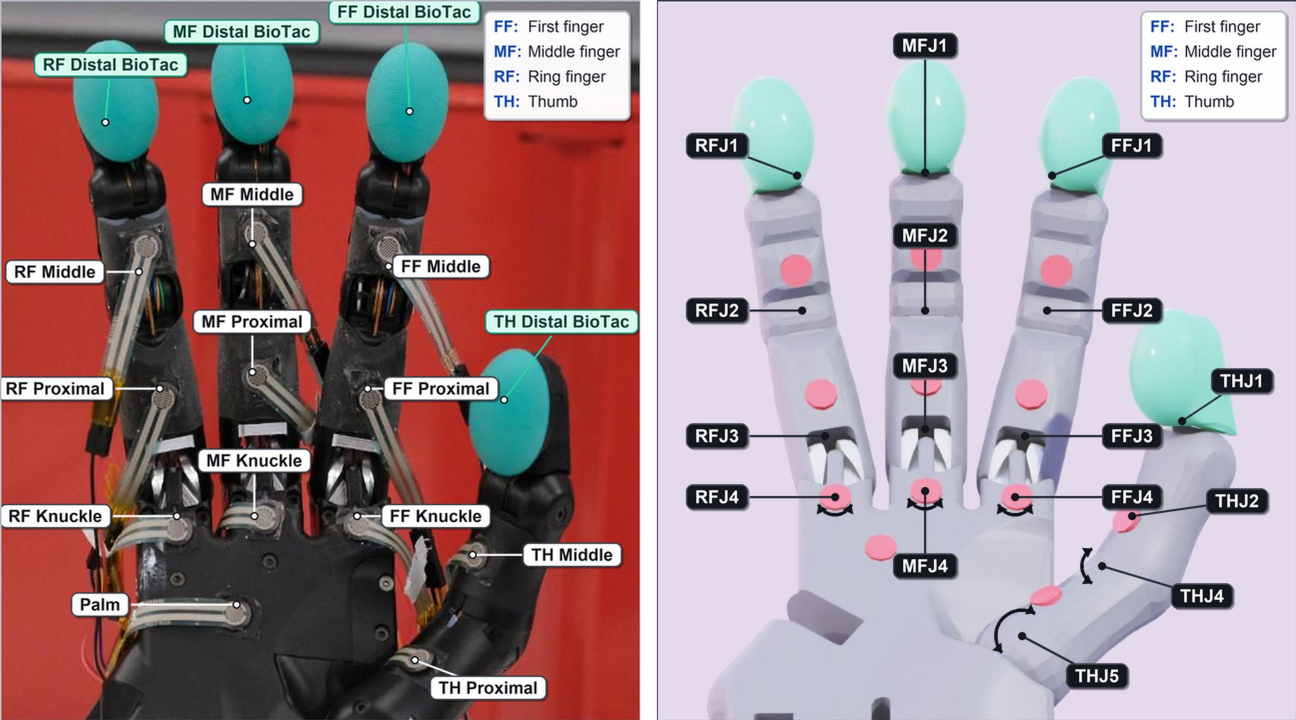}
    \caption{\textbf{FSR setup on the Shadow Dexterous Hand Lite.} (\textbf{a}) The physical hand, with the FSR placements labelled. (\textbf{b}) The simulation model of the same hand, with the joints labelled; the FSRs are shown in red.}
    \label{fig:physical_hand_sim_hand}
\end{figure}


\section{METHODOLOGY}


\subsection{Problem Setting}

The aim of this work is to record sustained Baoding ball rotations on the Shadow Dexterous Hand Lite through zero-shot sim-to-real transfer of an RL policy. Two identical smooth Baoding balls ($1.58$~inch in diameter, mass $55$\,g) are rotated
continuously through finger-based contacts, with the episode ending when a ball leaves the hand. 
We formulate the task as a Partially Observable Markov Decision Process (POMDP) defined by the tuple $(S,A,T,O,R,\gamma)$.
We note here that the simulated transition dynamics $T$ is an idealisation of the real-world: the physical hand does not reach the
same state under the same command because of the approximate tendon model, actuator
dynamics, and hand--ball and ball--ball contact. Likewise, $O$ is idealized, since hardware proprioception carries calibration offsets, while the tactile channel exhibits latency and intermittent activations, so the same physical state can produce a different observation. A policy optimized under $(T,O)$ therefore is not guaranteed to reproduce that behaviour on hardware, which is the exact problem we seek to mitigate.

\subsection{Agent and Learning Setup}
\label{sec:setup}

The agent is blind. Its observation consists of a four-frame stack, where each frame
contains 52 proprioceptive values, comprising joint position, joint velocity,
position error, and previous action for each of the 13 actuators, together with 16
binary tactile values, giving a 272-dimensional observation. No camera, depth, motion
capture, or privileged state is available at inference. The action is a 13-dimensional joint-position command issued at 60~Hz. The same
observation and action layout is used in simulation and on hardware. The hand is held at an approximately $15^\circ$ tilt throughout, and the two balls
are seated diagonally opposed at the start of every episode.
We retain the reward from RoTO~\cite{miller2026enhancing} unchanged, which consists of
a dense term that encourages the hand to keep the balls from being dropped and the sparse term rewards a completed exchange when both balls reach their respective goal markers.
Policies are trained with PPO in Isaac Lab using 10~s episodes, corresponding to 600
 control steps. 
 

\subsection{Hardware Platform and Sensing}
\label{sec:coupling}

The hardware system used in the study is a Shadow Dexterous Hand Lite with 16 kinematic joints (Figure~\ref{fig:physical_hand_sim_hand}). It comprises three fingers with three joints each and a four-joint thumb. The proximal interphalangeal (PIP) and distal interphalangeal (DIP) joints of the index, middle, and ring fingers are mechanically coupled, resulting in 13 independently commandable actuators.

The tactile sensing system uses twelve force-sensitive resistors (FSRs) distributed across the palm and phalanges, together with four BioTac sensors on the fingertips, at the DIP joints of the index, middle and ring fingers and at the thumb interphalangeal (IP) joint
(Figure~\ref{fig:physical_hand_sim_hand}). The sites are chosen from per-link contact frequencies observed during a simulation run, with the metacarpophalangeal (MCP) regions having the highest frequency, followed by the PIP regions and the palm. The 16 tactile channels are converted into binary contact signals using a thresholding mechanism described in Section~\ref{cal}.

\subsection{Simulation Embodiment}
\label{sim embody}

\begin{figure}[t]
    \centering
    \includegraphics[width=0.75\columnwidth]{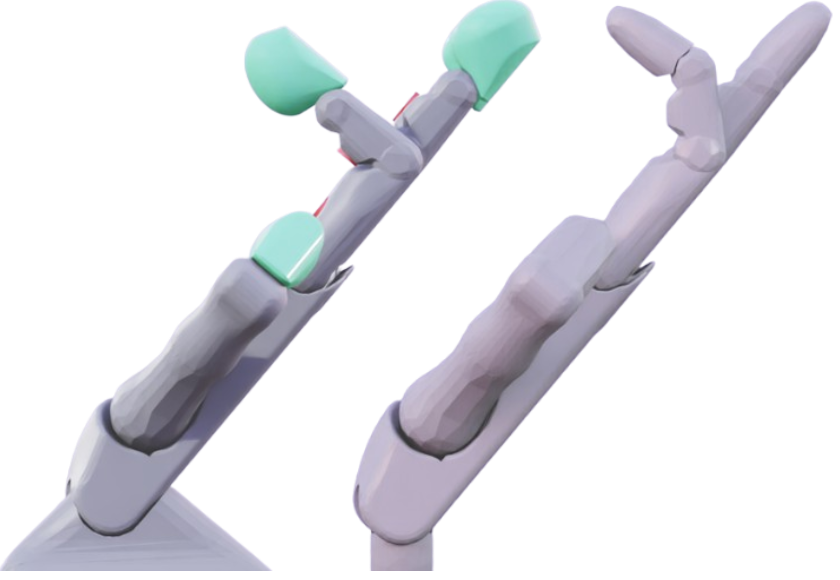}
    \caption{\textbf{Fixed tendon coupling on the Shadow Hand.} The same $45^\circ$ PIP flexion command applied to both simulation embodiments (\textbf{a}): the coupling mechanism used in this work, fitted with BioTac fingertips. (\textbf{b}): the coupling mechanism as supplied in Isaac Lab.}
    \label{fig:coupling}
\end{figure}

We observed that the default coupling in the Isaac Lab Shadow Dexterous Hand asset is inconsistent with the coupling mechanism of the physical Shadow Dexterous Hand (Fig.~\ref{fig:coupling}). On hardware, the coupled joints move sequentially: during flexion, the PIP joint moves first while the DIP joint remains near zero until the PIP joint reaches approximately $100^\circ$, after which the DIP joint begins to move. During extension, the order is reversed, with the DIP joint returning before the PIP joint moves. In contrast, the default Isaac Lab asset models the DIP angle as a fixed multiple of the PIP angle, causing both joints to move simultaneously. Consequently, the default asset cannot be used directly as a hardware-aligned model.

We therefore implement the hardware coupling explicitly at each environment step. For each pair of coupled joints, the policy outputs a single command, and the command is applied sequentially according to the current joint configuration: during flexion, the PIP joint is actuated until the $100^\circ$ coupling threshold is reached before the DIP joint is actuated, while during extension the order is reversed. For tactile sensing, small collision bodies are added at the sensor locations in simulation to match the coverage and location of the physical FSR sensors. Fig.~\ref{fig:physical_hand_sim_hand} shows the physical hand with the FSR sensors alongside the corresponding simulated embodiment.

\subsection{Bridging the Sim-to-Real Gap}
\begin{figure*}[t!]
  \centering
  \includegraphics[width=\linewidth]{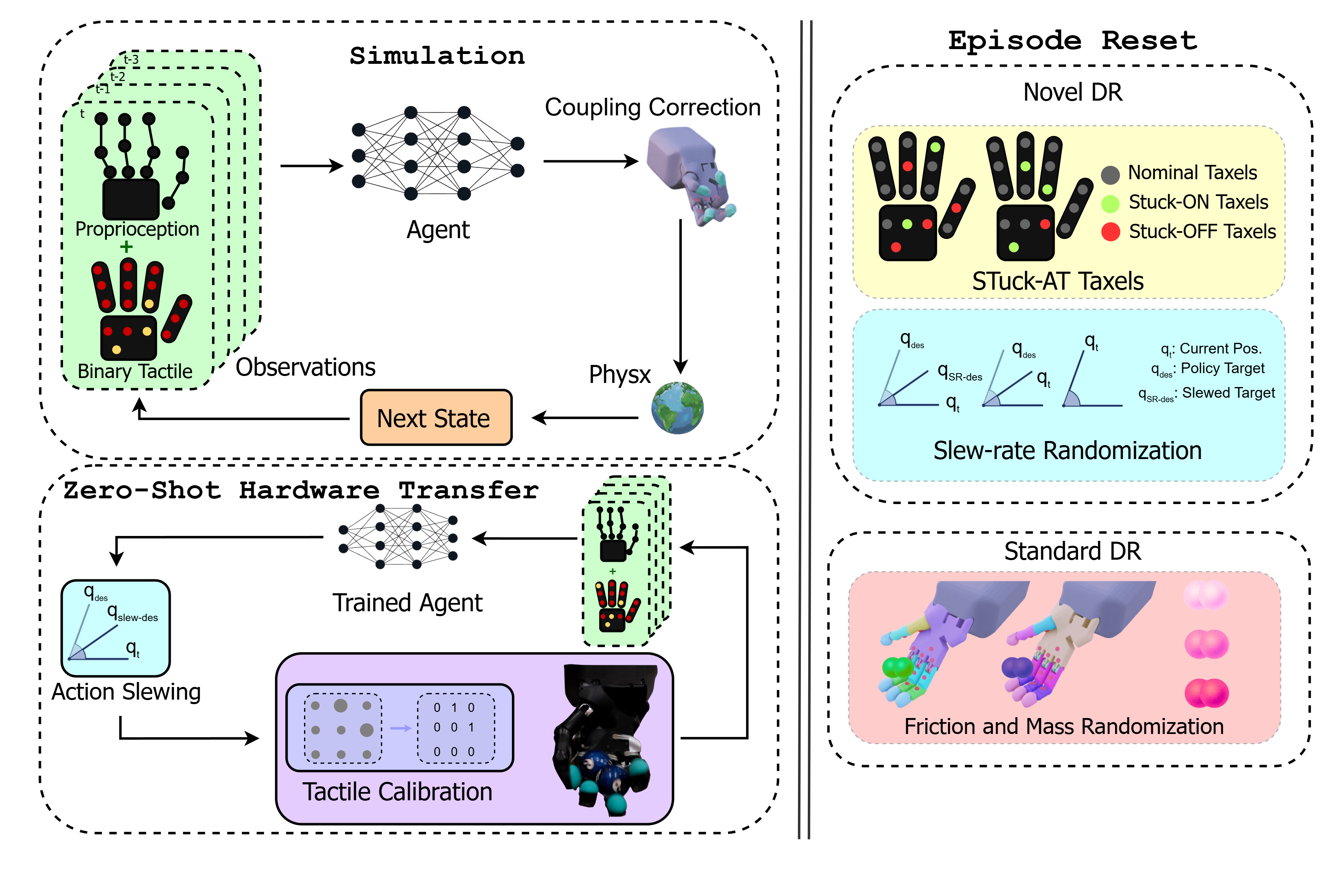}
  \caption{\textbf{Method.} (\textbf{Top left}) Policy training in simulation. The agent's actions are passed through the corrected coupling mechanism before the physics step, and the resulting state is returned as proprioceptive and binary tactile observations. (\textbf{Bottom left}) Zero-shot deployment on hardware at 60~Hz.. The trained policy is transferred without modification, with tactile calibration and actions rate-limited during execution.  (\textbf{Right}) Randomization applied at each episode reset, including the STAT and SlewDR terms introduced in this work, together with the standard ball-mass and friction randomization.
}
  \label{fig:method}
\end{figure*}

After aligning the simulated embodiment with the corrected sequential coupling and tactile sensing sites, residual gaps remain in the contact dynamics, joint motions and tactile
thresholding. These are addressed by the following complementary methods, illustrated in Figure~\ref{fig:method}. We introduce Slew-rate Domain Randomisation (SlewDR) and Stuck-AT Taxels (STAT) to address these discrepancies, together with tactile calibration for zero-shot deployment on hardware.

\subsubsection{Physical Parameter Randomisation} Physical parameters are resampled at the start of each training episode to expose the policy to variation in ball and hand contact dynamics. Ball mass is sampled from $45$--$100\,\mathrm{g}$, while static and dynamic friction are randomised independently across the balls and the distal, middle, proximal, knuckle, and palm regions of the hand. The friction ranges are selected to reflect the differing surface properties of these regions on the physical hand, with higher friction assigned to the distal segments to account for the silicone skin covering the BioTac fingertip sensors and lower values used for the smoother palm surface.

\subsubsection{Slew-rate Domain Randomisation (SlewDR)}
Even with soft gains in simulation and equally low gains on the real hand, the policy still commanded joint motion faster than the hardware could execute. We suspect this is due to tendon friction and reflected motor inertia, which the simulated joint does not represent. We therefore apply a rate limit to actions at deployment, constraining the change in commanded position between consecutive steps. The same limit is applied in simulation as a randomisation, with the scale factor $s\sim\mathcal{U}(0.3,1.0)$ resampled at each episode reset. The published joint target is then


\begin{equation}
\begin{split}
  q^{\mathrm{SR\text{-}des}}_{t}
  =
  q_{t}
  + \operatorname{clip}\big(&
    q^{\mathrm{des}}_{t}-q_{t},\\
    &-s\,\dot{q}^{\max}\Delta t,\;
    s\,\dot{q}^{\max}\Delta t\big),
\end{split}
\label{eq:slew}
\end{equation}

where $\dot{q}^{\max}$ is the rated joint velocity limit, $\Delta t=1/60$\,s,  $q_{\mathrm{des},t}$ denotes the desired joint position and $q_{t}$ denotes the joint-position command actually published to the robot. 
The same rate limit defined in Equation~\eqref{eq:slew} is applied on hardware. Unlike
in simulation, the scale factor $s$ is not resampled: it is held fixed throughout an
episode and remains at the same value across every deployment episode. The values used
are reported in the Section \ref{eval-proto}.

\subsubsection{Stuck-AT Taxels (STAT)}
Policies trained on clean tactile input were found to over-rely on individual
channels, which negatively impacted sim-to-real when these channels did not provide reliable readings. To counter this, we propose a novel domain randomisation technique called Stuck-AT Taxels. We corrupt the tactile channels during training, in the manner of
dropout, so that no single channel can be treated as reliable. At each episode reset, a subset of $k$ of the twelve FSR channels is uniformly selected at random, and each selected channel $i$ is held at the fixed value $v_i$ for the whole episode, while the remaining channels report the true simulated contact state. 

\begin{equation}
  k \sim \mathcal{U}\{0,\ldots,6\}, \qquad v_i \sim \mathcal{U}\{0,1\},
  \label{eq:tactile-corruption}
\end{equation}


\subsubsection{Tactile Calibration}
\label{cal}
We observed that using static FSR thresholds was insufficient for effective sim-to-real transfer, as the threshold value changed with the hands static pose, wire movement and the corresponding motion during deployment, even with no object in the hand. Thus, we converted each FSR channel into a binary contact value with 2 thresholds (high and low) to provide hysteresis:
\begin{align}
  \tau^{\mathrm{hi}}
    &= P_{99.5}(x) + 5,
    \label{eq:calib-hi} \\[6pt]
  \tau^{\mathrm{lo}}
    &=
    \begin{cases}
      \mathrm{med}(x)+2,
        & \text{if } P_{99.5}(x)-\mathrm{med}(x) \le 15, \\[4pt]
      \tau^{\mathrm{hi}}-3,
        & \text{otherwise.}
    \end{cases}
    \label{eq:calib-lo}
\end{align}
where $P_{99.5}(x)$ is the $99.5$th percentile and $\mathrm{med}(x)$ is the median of the robust baseline value calculated from empty-hand FSR readings before deployment.
The additives $+5$, $+2$, and $-3$ are chosen empirically for this specific sensor layout. Binarization of the contact signal follows a Schmitt rule, it switches on above $\tau^{\mathrm{hi}}$ and switches off below $\tau^{\mathrm{lo}}$, otherwise retains the previous timestep's state.



\section{Experimental Setup}




\subsection{Evaluation Protocol}
\label{eval-proto}

We evaluated the trained policies on the physical Shadow Dexterous Hand Lite, mounted at approximately $15^\circ$ downward from horizontal, with control commands issued at 60~Hz. The default proportional and derivative gains provided for the Shadow Dexterous hand Lite were scaled after testing by $0.15$, except for the MCP flexion joints of the index, middle and ring fingers, which were kept at default, and the MCP abduction joints of the same three fingers, which were scaled by $0.3$. These PD gains were kept fixed across all evaluated configurations.

All policies were deployed zero-shot, without hardware training, policy fine-tuning, or checkpoint selection based on hardware performance. The same pair of $1.58$-in, 55-g Baoding balls was used across all hardware trials. For each configuration (Table~\ref{tab:configurations}), 10 independent trials were performed, with the balls placed by hand before each trial and not disturbed thereafter. Trials ran until a ball left the hand, with no time limit imposed. The command slew scale $s$ was held fixed within and across trials, with $s=0.65$ for the proprioception-only policy and $s=0.53$ for all other policies.
In simulation, each configuration was evaluated over 768 episodes using 256 parallel environments and three training seeds. Episodes were limited to 10~s, with policy control at 60~Hz and physics simulation at 240~Hz. The evaluation used the same balls with the same initial-position perturbation ($\pm5$~mm spawn noise) to ensure that none of the evaluation environments were fully deterministic.

\begin{table}[b]
    \centering
    \caption{Evaluated configurations}
    \label{tab:configurations}
    \resizebox{\columnwidth}{!}{%
    \begin{tabular}{lcc}
        \toprule
        Configuration & Training Randomization & Feedback / Input \\
        \midrule
        C1
            & Physical interaction
            & Proprioception + tactile \\
        C2
            & Physical + SlewDR
            & Proprioception + tactile \\
        \textbf{C3 (ours)}
            & \textbf{Physical + SlewDR + STAT}
            & \textbf{Proprioception + tactile} \\
        Proprio-only
            & Physical interaction
            & Proprioception only \\
        C3-TacOff
            & C3
            & Proprioception only$^\dagger$ \\
        Open-loop
            & --
            & None (recorded trajectory) \\
        \bottomrule
    \end{tabular}%
    }
    
    \raggedright
    \footnotesize
    $^\dagger$ C3-TacOff uses the C3 checkpoint with tactile observations zeroed at deployment.
\end{table}

\subsection{Evaluation Metrics}


\subsubsection{Real-world Metrics}
\begin{itemize}
\item \textbf{Full Rotations:} the total number of complete $360^\circ$ rotations completed before a ball leaves the hand. A full rotation consists of two consecutive successful ball exchanges. Each exchange transfers the pair of balls around the hand and swaps their relative positions; after two exchanges, the pair returns to its original relative configuration, corresponding to one complete $360^\circ$ rotation.

\item \textbf{Time-to-drop ($t_{\mathrm{drop}}$):} the elapsed time from the the moment both balls are seated in the hand until either ball leaves the hand.

\item \textbf{Rotational speed:} the rate of rotation
$R_{10}/\min(10,t_{\mathrm{drop}})$ (rot/s), where $R_{10}$ is the number of
complete rotations within the first $10~s$.

\item \textbf{Exchange efficiency ($\eta$):} the fraction of executed gait cycles
that result in a completed ball exchange, computed as
\begin{equation}
    \eta = \frac{2R_{10}}{f_{\mathrm{gait}}T},
\end{equation}

where $f_{\mathrm{gait}}$ is the gait frequency measured from the joint trace and
$T=\min(10,t_{\mathrm{drop}})$. Since at most one exchange can occur per gait
cycle, $\eta\leq1$.

\end{itemize}

\subsubsection{Simulation-to-real Metrics}
Simulation episodes are limited to 10\,s. We therefore compare domains using
rotational speed over the same early window: $R_{10}/\min(10,t_{\mathrm{surv}})$ in simulation, where
$t_{\mathrm{surv}}$ is episode survival time in seconds. Alongside it we report the \textbf{Drop Rate}, the fraction of episodes in which either ball leaves the hand within the 10\,s.

For each configuration, hardware results are reported as mean $\pm$ standard
deviation over 10 trials. Simulation results are aggregated over 768 episodes across
three training seeds.

\section{Results}
\label{sec:results}

\subsection{Real-world Performance}

\begin{table*}[t]
\caption{Hardware results are mean$\pm$std over 10 trials; simulation results are aggregated over 768 episodes. Dashes ({---}) denote metrics not evaluated or reported.
}
\vspace{-2em}
\label{tab:main_results}
\begin{center}
\small
\resizebox{\textwidth}{!}{%
\begin{tabular}{@{}l cc cc c cc c cc@{}}
\toprule
& \multicolumn{2}{c}{\textbf{Full rotations}}
& \multicolumn{2}{c}{\textbf{Time to drop (s)}}
& \textbf{$\eta$}
& \multicolumn{2}{c}{\textbf{Speed (HW)}}
& \textbf{Speed (sim)}
& \multicolumn{2}{c}{\textbf{Drop rate}} \\
\cmidrule(lr){2-3}\cmidrule(lr){4-5}\cmidrule(lr){6-6}\cmidrule(lr){7-8}\cmidrule(lr){9-9}\cmidrule(lr){10-11}
\textbf{Condition}
& mean $\pm$ std & min--max
& mean $\pm$ std & min--max
& mean $\pm$ std
& mean $\pm$ std & min--max
& mean $\pm$ std
& sim & HW \\
\midrule
\multicolumn{11}{@{}l}{\textit{\textbf{Prior hardware results:}}} \\
Nagabandi et al.~\cite{nagabandi2020deep}$^\dagger$
  & $\sim 0.50$ & {---}
  & {---} & {---}
  & {---}
  & {---} & {---}
  & {---}
  & {---} & {---} \\
Yuan et al.~\cite{yuan2024robot}$^\ddagger$
  & $\sim 3.00$ & {---}
  & {---} & {---}
  & {---}
  & {$\sim$0.30} & {---}
  & {---}
  & {---} & {---} \\
Yang et al.~\cite{yang2023tacgnn}$^\S$
  & $\sim 0.50$ & {---}
  & {---} & {---}
  & {---}
  & {---} & {---}
  & {---}
  & {---} & {---} \\
\midrule
\multicolumn{11}{@{}l}{\textit{\textbf{Benchmark \& ablations:}}} \\
Open-loop
  & $1.7 \pm 1.99$ & 0--5.0
  & $6.1 \pm 4.7$   & 1--14
  & {---}
  & $0.168 \pm 0.179$ & 0--0.4
  & {---}
  & {---} & 80\% \\
Prop.-only
  & $4.4 \pm 2.44$ & 0.5--8.0
  & $21.4 \pm 12.2$ & 4--43
  & $0.31 \pm 0.13$
  & $0.226 \pm 0.098$ & 0.05--0.35
  & $0.591 \pm 0.128$
  & 17\% & 20\% \\
C3-TacOff
  & $5.0 \pm 4.1$ & 0--14.5
  & $19.0 \pm 11.2$ & 3--40
  & $0.58 \pm 0.26$
  & $0.256 \pm 0.112$ & 0--0.4
  & $0.291 \pm 0.167$
  & 32\% & 20\% \\
\midrule
\rowcolor{green!12}
\textbf{C3 (ours)}
  & $\mathbf{34.75 \pm 26.06}$ & \textbf{2.0--74}
  & $\mathbf{94.8 \pm 68.0}$   & \textbf{8--201}
  & $\mathbf{0.9 \pm 0.16}$
  & $\mathbf{0.36 \pm 0.061}$ & \textbf{0.25--0.45}
  & $\mathbf{0.416 \pm 0.129}$
  & \textbf{2\%} & \textbf{10\%} \\
\midrule
\rowcolor{blue!10}
C1
  & $0.35 \pm 0.63$ & 0--2
  & $5.9 \pm 4.3$   & 2--13
  & {---}
  & $0.054 \pm 0.079$ & 0--0.2
  & $0.959 \pm 0.071$
  & 2\% & 70\% \\
\rowcolor{orange!15}
C2
  & $0.5 \pm 0.53$ & 0--1.5
  & $4.1 \pm 1.5$   & 2--7
  & {---}
  & $0.101 \pm 0.094$ & 0--0.25
  & $0.277 \pm 0.082$
  & 9\% & 100\% \\
\bottomrule
\end{tabular}%
}
\end{center}
{\footnotesize
$^\dagger$Nagabandi et al.~\cite{nagabandi2020deep} report a $\sim$54\% success rate at completing a single $180^\circ$ half-rotation within a fixed 10\,s horizon (interpreted as $\sim$0.5 full rotations on successful attempts; inferred from video demonstration); multi-trial variance, time-to-drop, speed, and continuous rotations were not evaluated. \par
$^\ddagger$Yuan et al.~\cite{yuan2024robot} report $\sim$3 full rotations achieved within a fixed 10\,s horizon ($\sim$0.30\,rot/s); trial-to-trial variance, time-to-drop distributions, and continuous unbounded longevity were not reported. \par
$^\S$Yang et al.~\cite{yang2023tacgnn} demonstrate a single qualitative $180^\circ$ half-rotation execution on physical hardware without object vision, but report no quantitative continuous trial distributions, speed, or drop metrics.
}
\end{table*}

\begin{figure}[t]
    \centering
    \includegraphics[width=1\columnwidth]{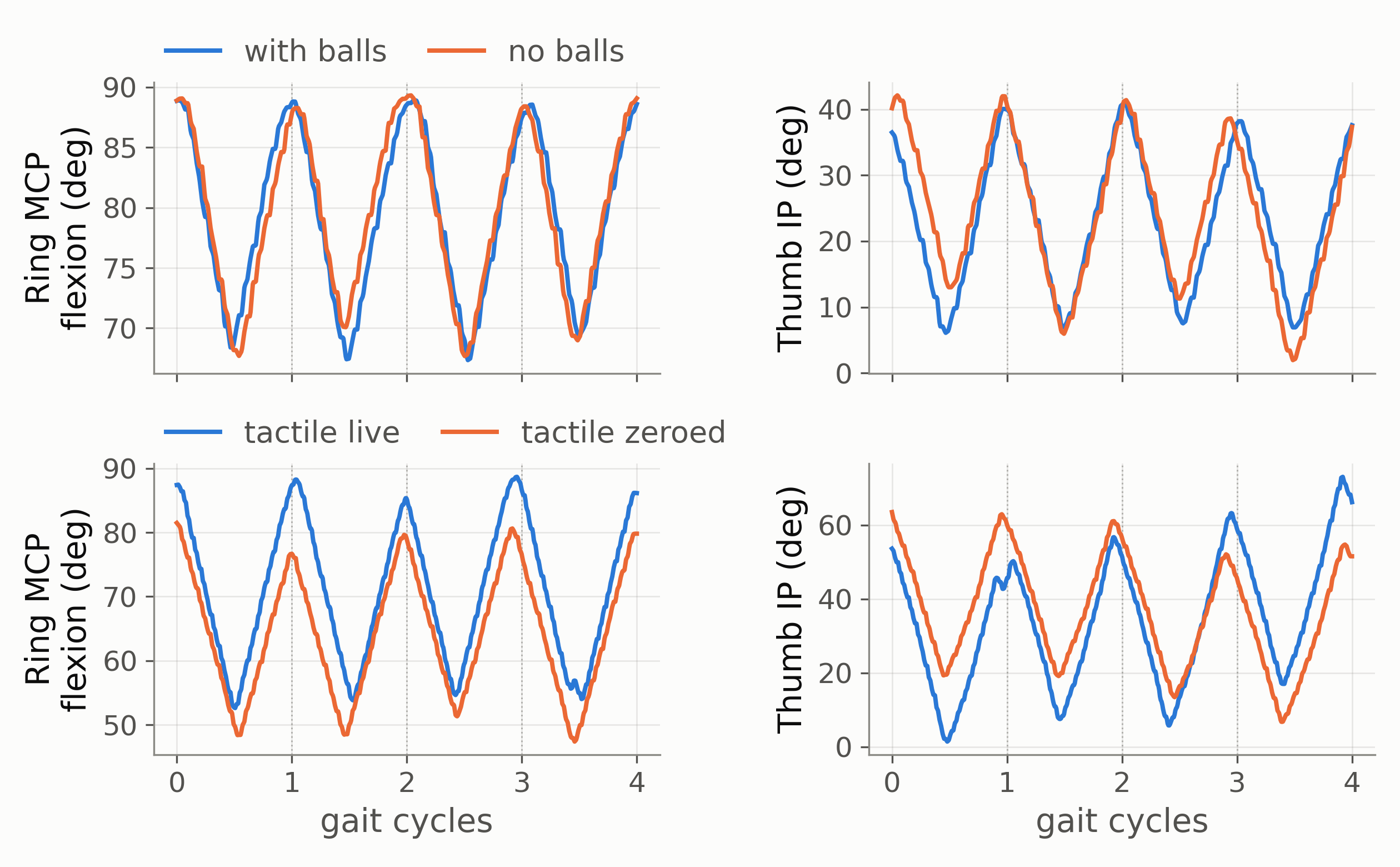}
    \caption{Joint trajectories for the ring-finger MCP flexion joint and the thumb IP joint over four rotation cycles. Top row: the proprioception-only policy with and without the balls in the hand. Bottom row: C3-TacOff against C3. Each plot overlays the two conditions for a single joint.}
    \label{fig:prop_only_and_Zero_tac-C3}
\end{figure}

The evaluated configurations and their corresponding training randomization and feedback settings are summarized in Table~\ref{tab:configurations}.
The C3 policy leads on all three metrics, achieving $34.75\pm26.06$ full rotations, a time-to-drop of $94.8\pm68.0$~s, and a rotation speed of $0.360\pm0.061$ across 10 trials. All trials reached at least two full rotations, and 9/10 continued beyond the 10-s window. Removing tactile feedback from the same policy in C3-TacOff reduced performance from 34.75 to 5.00 full rotations and from 94.8~s to 19.0 time-to-drop, while the corresponding decrement in rotation speed was smaller, from 0.36 to 0.256. The separately trained proprioception-only policy achieved 4.40 rotations compared with 1.70 for open-loop replay, while its performance was close to C3-TacOff (4.40 vs.\ 5.00 rotations). 


The effect of the cumulative randomization stack is evaluated through the
C1--C3 ladder as shown in Table~\ref{tab:main_results}. C3 substantially outperforms the
two intermediate configurations across all reported hardware metrics. C1 and C2 both fail to sustain the task on hardware, with median performance of 0.00 and 0.50 full rotations and median times-to-drop below 6~s. The transition from C1 to C2 provides no measurable gain: full rotations increase only from 0.35 to 0.50 with overlapping ranges, while time-to-drop decreases from 5.9 to 4.1~s. C3, by contrast, increases from 0.50 to 34.75 full rotations and from 4.1 to 94.8~s in time-to-drop relative to C2. The low variance of C1 and C2 reflects consistent failure.


\subsection{Simulation-to-real Comparison}

The simulation and hardware results for the common evaluation metrics and the randomization ladder are shown in Table~\ref{tab:main_results}. C3 and C3-TacOff show the closest simulation-to-hardware agreement among all configurations in terms of rotation speed.

\subsection{Long-Horizon Demonstration}
A separate deployment of the C3 policy completed 112.5 full $360^\circ$ rotations in 310 seconds.  The run used the same checkpoint and deployment conditions
as the 10-trial evaluation, but was conducted outside that set and is therefore not
included in the reported statistics.

\section{Discussion}

\subsection{What is tactile providing?}

We evaluate four policies on hardware under different feedback conditions: open-loop replay, proprioception-only, C3-TacOff, and C3. 
\textbf{Open-loop.} Open-loop replay was the least robust condition (Table~\ref{tab:main_results}), as the fixed trajectory cannot adapt and is highly sensitive to the initial ball configuration.
\textbf{Proprioception-only.} The separately trained proprioception-only policy produced the results shown in Table~\ref{tab:main_results}. It exhibited one of the fastest gaits among the four conditions while achieving one of the lowest exchange efficiencies. This mismatch between fast motion and low efficiency suggests that a faster gait does not necessarily result in more successful rotations. 
Removing the balls had little qualitative effect on the policy behaviour, as shown in Fig.~\ref{fig:prop_only_and_Zero_tac-C3}, where the joint trajectories remain largely unchanged across gait cycles with and without the balls. Similar behaviour was observed when the position-error and velocity inputs were zeroed in simulation. These observations suggest that the policy primarily executes a repeated learned cycle rather than continuously adapting its motion to the instantaneous object configuration, consistent with its low exchange efficiency.


\textbf{C3-TacOff.} We use the C3 checkpoint with tactile observations zeroed at deployment. C3-TacOff achieved the results shown in Table~\ref{tab:main_results}, with performance very similar to the separately trained proprioception-only policy across the evaluated hardware metrics. It maintained a similar gait frequency to C3, completing approximately 8.1 versus 7.9 cycles, but with substantially lower exchange efficiency. Qualitatively, C3-TacOff exhibited a repeated gait similar to the proprioception-only policy, with balls moving between neighbouring fingers until aligned for exchange. Its gait speed was also similar between simulation and hardware, while the simulation drop rate was significantly higher when tactile feedback was removed. This suggests that C3-TacOff retains the learned gait but loses the tactile-based corrections needed to sustain the manipulation.


\textbf{C3.} With tactile input restored, C3 achieved the results shown in Table~\ref{tab:main_results}. The exchange efficiency of $\eta=0.90$ is close to the theoretical upper bound of one exchange per gait cycle. Compared to C3-TacOff the difference in sustained performance is substantial: C3 achieved approximately $7\times$ as many rotations and remained active for nearly $5\times$ as long before dropping. This difference in motion adaptation can also be seen in Fig.~\ref{fig:prop_only_and_Zero_tac-C3}. The joint trajectories of C3 show greater variation across gait cycles, with varying peak amplitudes and secondary peaks, whereas proprioception-only and C3-TacOff remain more repetitive. This indicates that C3 adjusts its motion across cycles rather than simply repeating the nominal gait. Comparing C3 and C3-TacOff in simulation, the drop rate decreases from 32\% to approximately 2\% with tactile input. This is consistent with the behaviour observed on hardware, where tactile feedback enables recovery from unfavorable ball configurations. In one observed recovery, the thumb initially failed to push one ball sufficiently toward the ring finger and instead retained the incoming ball, while the index and middle fingers repositioned the other ball. Once the required configuration was reached, the learned gait resumed.

Taken together, these results suggest \textbf{two complementary roles} for tactile feedback during in-hand Baoding ball rotation. First, tactile feedback improves \textbf{\textit{exchange completion}}, allowing a larger fraction of the executed gait cycles to result in successful exchanges. Although the learned gait can be executed without tactile feedback, \textbf{many of these cycles do not lead to a successful exchange}. Second, tactile feedback improves \textbf{\textit{task sustainment}} by allowing the policy to recover from unfavorable ball configurations. When the balls deviate from the configuration required by the nominal gait, tactile feedback enables the policy to adjust its motion and \textbf{restore the intended configuration to continue the manipulation}.

\begin{figure}[t]
    \centering
    \includegraphics[width=0.8\columnwidth]{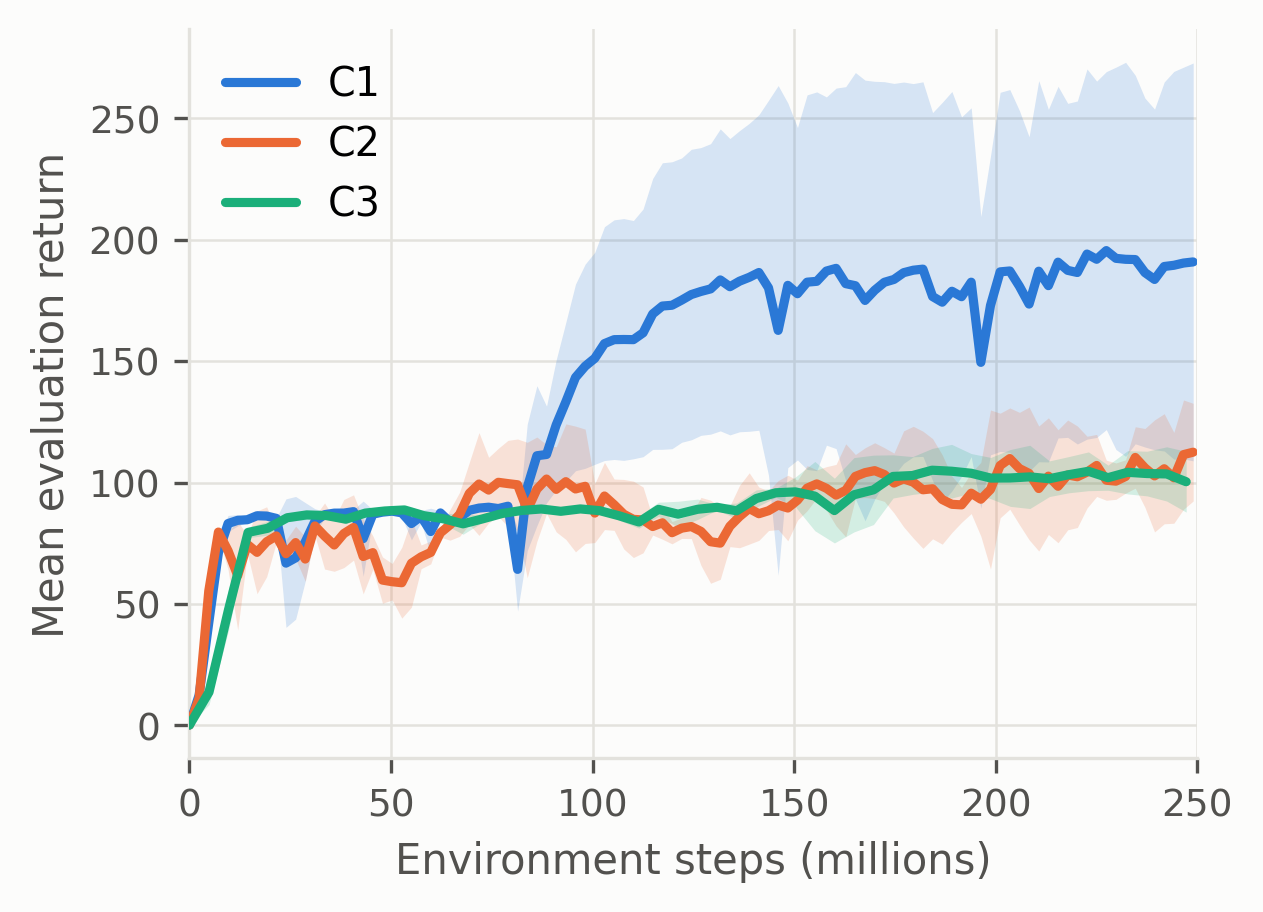}
    \caption{Mean evaluation return during training for configurations C1, C2, and C3, each run across three seeds for 250 M environment steps. The 100 mark corresponds approximately to retaining both balls without rotating them, so returns above it indicate that rotations have begun. C1 crosses it at around 80 M steps against approximately 160–170 M for C2 and C3, and reaches roughly twice their final return.}
    \label{fig:seeded_returns}
\end{figure}

\begin{figure}[t]
    \centering
    \includegraphics[width=1\columnwidth]{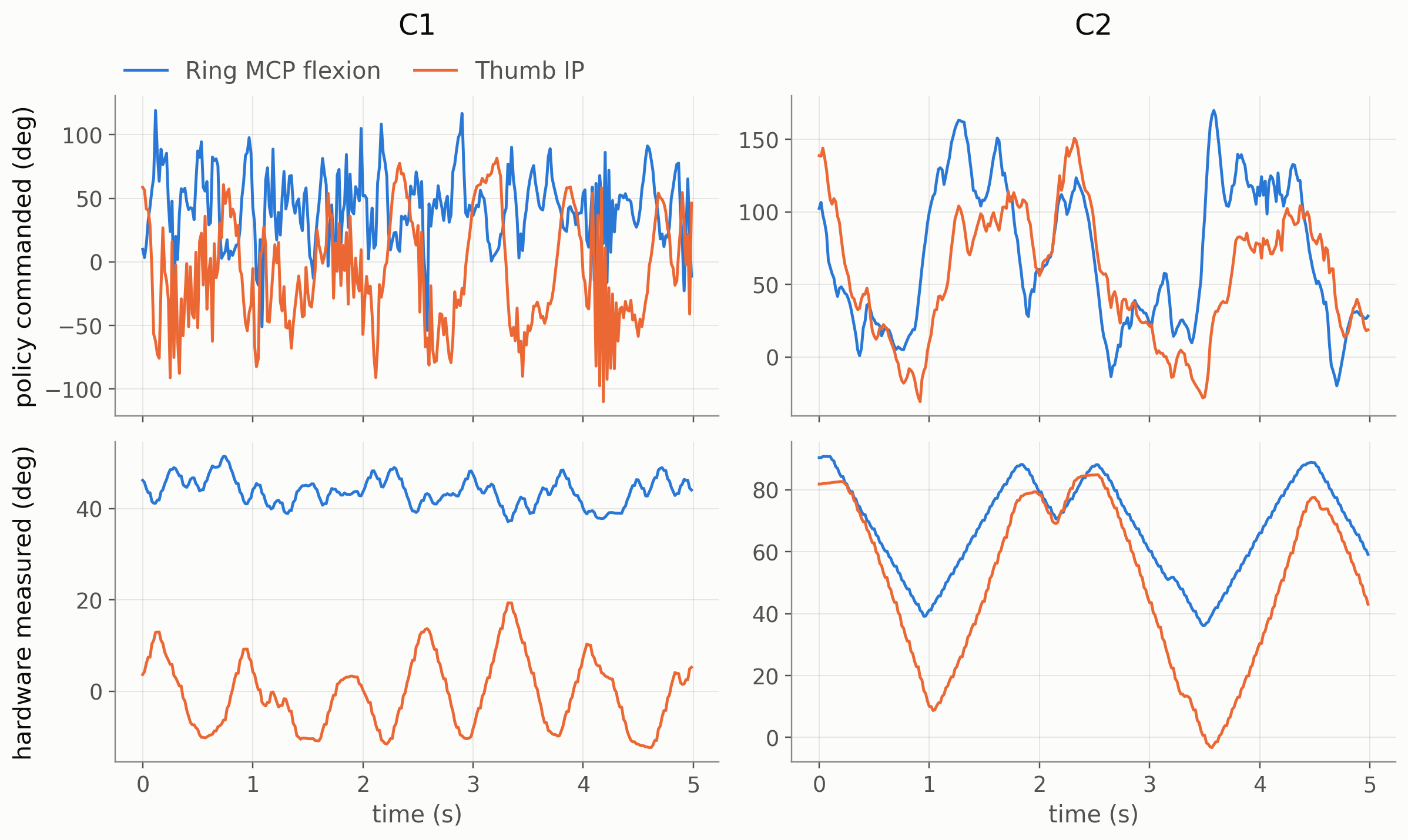}
    \caption{Commanded and achieved joint positions on hardware for two configurations. Left column: C1. Right column: C2. Top row: the positions commanded by the policy. Bottom row: the positions measured on the physical hand over the same interval. Each panel plots two joints: the ring-finger MCP flexion joint and the thumb IP joint.}
    \label{fig:sim_hw_actions}
\end{figure}

\subsection{How do these randomization enable transfer? }

C1 performs the worst on hardware despite the best simulation performance (Table~\ref{tab:main_results}). The simulated gait approaches one rotation per second, but this aggressive motion does not transfer to hardware, where execution is substantially slower and difficult to sustain. The difference is also visible in Fig.~\ref{fig:sim_hw_actions}, where the simulated actions are larger and more rapid than those observed on hardware.

\begin{figure}[b]
    \centering
    \includegraphics[width=1\columnwidth]{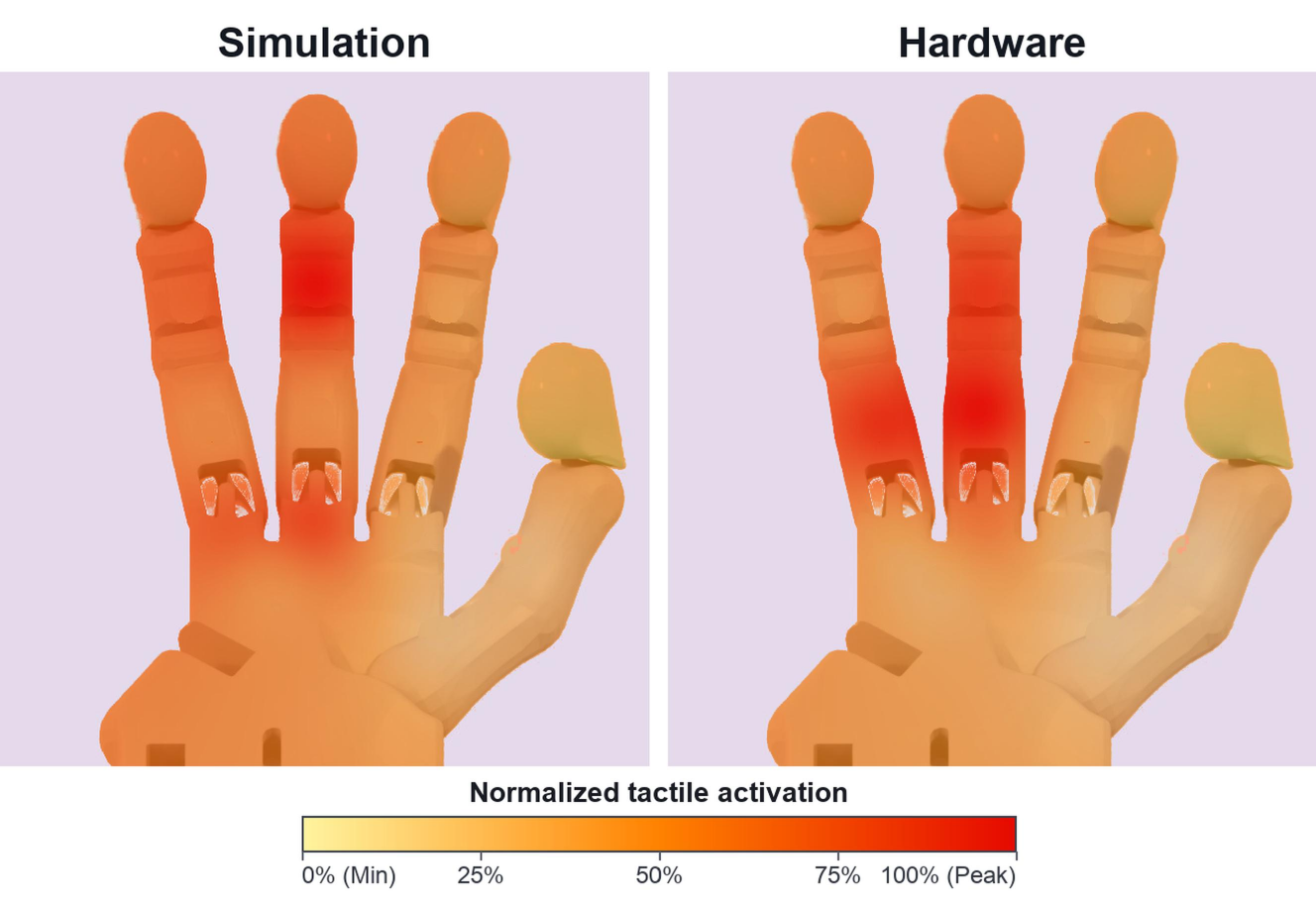}
    \caption{Normalised binary-tactile activation for C3 policy in \textbf{simulation (left) and hardware (right)}. In simulation, activation is concentrated primarily around the middle and knuckle regions of the middle and ring fingers. On hardware, the dominant activation shifts proximally, with the proximal middle and ring-finger sensors accounting for a substantial amount of tactile activation. 
    }

    \label{fig:tacact}
\end{figure}
C2 introduces slew-rate randomization during simulation. The rate limit caps the change in commanded joint position at each step, so large policy outputs cannot produce equally large instantaneous motions. The policy therefore \textbf{learns to avoid abrupt commands and produce smoother motion that is easier for the hardware to track}, as shown in Fig.~\ref{fig:sim_hw_actions}, resulting in an increase in the mean number of rotations on hardware compared with C1 (Table~\ref{tab:main_results}). The simulated gait speed also drops substantially, reaching the lowest speed among the evaluated configurations, while the simulation and hardware speeds are correspondingly closer. However, introducing a range of slew rates makes the RL problem substantially harder, as reflected by the drop in reward across the three training seeds in Fig.~6. Although the motion transfers more closely to hardware, all hardware trials still ended in a drop.

\textbf{Addition of STAT.} STAT combines the smoother motion of C2 with substantially greater robustness, outperforming all previous policies across every measured hardware metric, as shown in Table~\ref{tab:main_results}. The remaining performance gap in C2 can be traced directly to its reliance on tactile feedback: when tactile input is removed from C2 in simulation, the drop rate rises to nearly 60\%, indicating that, despite its smoother gait, the policy remains heavily dependent on continuous tactile feedback to sustain the manipulation. STAT is trained specifically to reduce this dependence. When tactile observations are removed from the STAT-trained policy, the drop rate increases from 2.2\% to 32.2\%, substantially less than the degradation observed for C2, while the underlying motion remains active. This behaviour is consistent with the proprioception-only policy, which can execute the nominal gait without tactile feedback, albeit with substantially lower robustness.
Together, these results clarify the role of the randomization introduced by STAT. Rather than relying on tactile feedback to generate motion, STAT shifts the primary burden of execution onto the \textbf{learned gait itself}, with tactile feedback serving a supporting role in \textbf{local adjustment and correction}. \textbf{The policy therefore uses the learned gait as the primary driver of motion, while using tactile feedback to correct deviations in the ball configuration}, rather than relying on it continuously to sustain the motion.

\subsection{Which Taxels are important?}

To investigate where useful tactile information originates, we analysed the activation patterns of the 16 tactile channels in simulation and hardware (Fig.~\ref{fig:tacact}). Despite the availability of all channels, contact activations were concentrated around the middle and ring fingers, with comparatively limited activation around the index finger and palm. In simulation, the middle and MCP regions of these fingers exhibited the highest activation frequencies, whereas on hardware, activations shifted towards the proximal FSRs. 


Although the precise spatial distribution differed on hardware, activations concentrated around similar regions, suggesting that simulated contact frequency is a useful guide for sensor placement even if it does not predict exact channel-wise activation. Because only a few channels are strongly activated, such sim-to-real shifts would make a policy that depends on any individual channel sensitive to sensor faults or changes in contact density, which is addressed by STAT during training. Taken together, these observations suggest that \textbf{successful transfer depends less on which taxels are active than on how strongly the policy relies on any single one}.

\section{CONCLUSIONS}


We demonstrated zero-shot sim-to-real transfer of blind Baoding ball rotation on the Shadow Dexterous Hand Lite using only proprioception and binary tactile contact, without vision or any object-state estimation. Hardware-aligned coupling and sensor sites, hysteresis-based tactile calibration, and episode-level SlewDR and STAT narrow the sim-to-real gap enough for the policy to sustain rotation on the physical hand, completing 225 consecutive $180^\circ$ exchanges in a single continuous run. It is observed that the policy retains a rhythmic proprioceptive motion prior that continues without tactile input, while binary contact steers that prior into completed rotations and recovery. Open-loop replay, proprioception-only, and C3-TacOff ablations support this separation of roles and explain why removing touch collapses performance. A current limitation is the FSRs themselves: their contact latency, and their failure to register a smooth ball sliding quickly past, force a slower gait than the policy could otherwise sustain. Looking ahead, we aim to generalise across different ball sizes and masses, move beyond spherical objects to irregular shapes, and test low-latency sensors with discretised multi-level tactile readings, all under the same blind, zero-shot transfer setting.

\bibliographystyle{IEEEtran}
\bibliography{ref}

@article{andrychowicz2020learning,
  title={Learning dexterous in-hand manipulation},
  author={Andrychowicz, OpenAI: Marcin and Baker, Bowen and Chociej, Maciek and Jozefowicz, Rafal and McGrew, Bob and Pachocki, Jakub and Petron, Arthur and Plappert, Matthias and Powell, Glenn and Ray, Alex and others},
  journal={The International Journal of Robotics Research},
  volume={39},
  number={1},
  pages={3--20},
  year={2020},
  publisher={SAGE Publications Sage UK: London, England}
}

@inproceedings{akkaya2019solving,
  title={Solving rubik's cube with a robot hand},
  author={Akkaya, Ilge and Andrychowicz, Marcin and Chociej, Maciek and Litwin, Mateusz and McGrew, Bob and Petron, Arthur and Paino, Alex and Plappert, Matthias and Powell, Glenn and Ribas, Raphael and others},
  booktitle={2023 IEEE International Conference on Robotics and Automation (ICRA)},
  pages={5977--5984},
  year={2023},
  organization={IEEE}
}

@inproceedings{handa2023dextreme,
  title={Dextreme: Transfer of agile in-hand manipulation from simulation to reality},
  author={Handa, Ankur and Allshire, Arthur and Makoviychuk, Viktor and Petrenko, Aleksei and Singh, Ritvik and Liu, Jingzhou and Makoviichuk, Denys and Van Wyk, Karl and Zhurkevich, Alexander and Sundaralingam, Balakumar and others},
  booktitle={2023 IEEE International Conference on Robotics and Automation (ICRA)},
  pages={5977--5984},
  year={2023},
  organization={IEEE}
}

@article{chen2023visual,
  title={Visual dexterity: In-hand reorientation of novel and complex object shapes},
  author={Chen, Tao and Tippur, Megha and Wu, Siyang and Kumar, Vikash and Adelson, Edward and Agrawal, Pulkit},
  journal={Science Robotics},
  volume={8},
  number={84},
  pages={eadc9244},
  year={2023},
  publisher={American Association for the Advancement of Science}
}

@inproceedings{qi2023general,
  title={General in-hand object rotation with vision and touch},
  author={Qi, Haozhi and Yi, Brent and Suresh, Sudharshan and Lambeta, Mike and Ma, Yi and Calandra, Roberto and Malik, Jitendra},
  booktitle={Conference on Robot Learning},
  pages={2549--2564},
  year={2023},
  organization={PMLR}
}

@inproceedings{qi2023inhand,
  title={In-hand object rotation via rapid motor adaptation},
  author={Qi, Haozhi and Kumar, Ashish and Calandra, Roberto and Ma, Yi and Malik, Jitendra},
  booktitle={Conference on Robot Learning},
  pages={1722--1732},
  year={2023},
  organization={PMLR}
}

@inproceedings{sievers2022learning,
  title={Learning purely tactile in-hand manipulation with a torque-controlled hand},
  author={Sievers, Leon and Pitz, Johannes and B{\"a}uml, Berthold},
  booktitle={2022 International conference on robotics and automation (ICRA)},
  pages={2745--2751},
  year={2022},
  organization={IEEE}
}

@inproceedings{pitz2023dextrous,
  title={Dextrous tactile in-hand manipulation using a modular reinforcement learning architecture},
  author={Pitz, Johannes and R{\"o}stel, Lennart and Sievers, Leon and B{\"a}uml, Berthold},
  booktitle={2023 IEEE International Conference on Robotics and Automation (ICRA)},
  pages={1852--1858},
  year={2023},
  organization={IEEE}
}

@article{yang2023tacgnn,
  title={Tacgnn: Learning tactile-based in-hand manipulation with a blind robot using hierarchical graph neural network},
  author={Yang, Linhan and Huang, Bidan and Li, Qingbiao and Tsai, Ya-Yen and Lee, Wang Wei and Song, Chaoyang and Pan, Jia},
  journal={IEEE Robotics and Automation Letters},
  volume={8},
  number={6},
  pages={3605--3612},
  year={2023},
  publisher={IEEE}
}

@article{yang2024anyrotate,
  title={Anyrotate: Gravity-invariant in-hand object rotation with sim-to-real touch},
  author={Yang, Max and Lu, Chenghua and Church, Alex and Lin, Yijiong and Ford, Chris and Li, Haoran and Psomopoulou, Efi and Barton, David AW and Lepora, Nathan F},
  journal={arXiv preprint arXiv:2405.07391},
  year={2024}
}

@article{pan2026beyond,
  title={Beyond Binary: Sim-to-Real Dexterous Manipulation with Physics-Grounded Contact Representation},
  author={Pan, Jiahe and Coros, Stelian and Malik, Jitendra and Lin, Toru},
  journal={arXiv preprint arXiv:2605.28812},
  year={2026}
}

@inproceedings{chen2022system,
  title={A system for general in-hand object re-orientation},
  author={Chen, Tao and Xu, Jie and Agrawal, Pulkit},
  booktitle={Conference on robot learning},
  pages={297--307},
  year={2022},
  organization={PMLR}
}

@article{yin2023rotating,
  title={Rotating without seeing: Towards in-hand dexterity through touch},
  author={Yin, Zhao-Heng and Huang, Binghao and Qin, Yuzhe and Chen, Qifeng and Wang, Xiaolong},
  journal={arXiv preprint arXiv:2303.10880},
  year={2023}
}

@article{lee2024dextouch,
  title={Dextouch: Learning to seek and manipulate objects with tactile dexterity},
  author={Lee, Kang-Won and Qin, Yuzhe and Wang, Xiaolong and Lim, Soo-Chul},
  journal={IEEE Robotics and Automation Letters},
  volume={9},
  number={12},
  pages={10772--10779},
  year={2024},
  publisher={IEEE}
}

@inproceedings{nagabandi2020deep,
  title={Deep dynamics models for learning dexterous manipulation},
  author={Nagabandi, Anusha and Konolige, Kurt and Levine, Sergey and Kumar, Vikash},
  booktitle={Conference on robot learning},
  pages={1101--1112},
  year={2020},
  organization={PMLR}
}

@inproceedings{yuan2024robot,
  title={Robot synesthesia: In-hand manipulation with visuotactile sensing},
  author={Yuan, Ying and Che, Haichuan and Qin, Yuzhe and Huang, Binghao and Yin, Zhao-Heng and Lee, Kang-Won and Wu, Yi and Lim, Soo-Chul and Wang, Xiaolong},
  booktitle={2024 IEEE International Conference on Robotics and Automation (ICRA)},
  pages={6558--6565},
  year={2024},
  organization={IEEE}
}

@article{miller2026enhancing,
  title={Enhancing tactile-based reinforcement learning for robotic control},
  author={Miller, Elle and McInroe, Trevor and Abel, David and Mac Aodha, Oisin and Vijayakumar, Sethu},
  journal={Advances in Neural Information Processing Systems},
  volume={38},
  pages={129460--129494},
  year={2026}
}

@article{miller2026roto,
  title={roto 2.0: The Robot Tactile Olympiad},
  author={Miller, Elle and Reddy, Jayaram and Deshmukh, Ayush and McInroe, Trevor and Abel, David and Mac Aodha, Oisin and Vijayakumar, Sethu},
  journal={arXiv preprint arXiv:2605.21429},
  year={2026}
}

\end{document}